%% file: root.tex
\documentclass[letterpaper, 10 pt, conference]{ieeeconf}
\IEEEoverridecommandlockouts

\input{custom/packages}

\input{custom/acronyms}

\input{custom/commands}

\title{\LARGE \bf
Diffusion-Based Approximate MPC: \\ Fast and Consistent Imitation of Multi-Modal Action Distributions
}

\author{Pau Marquez Julbe$^{\dagger}$, Julian Nubert$^{\dagger,*}$, Henrik Hose$^{\ddagger}$, Sebastian Trimpe$^{\ddagger}$, and Katherine J. Kuchenbecker$^{\dagger}$
\thanks{This work is funded in part by the Max Planck ETH CLS and the German Research Foundation (DFG): RTG 2236/2 (UnRAVeL).\looseness=-1}
\thanks{$^{\dagger}$P.\ Marquez Julbe, J.\ Nubert, and K.\ J.\ Kuchenbecker are with the Max Planck Institute for Intelligent Systems, Stuttgart, Germany.\looseness=-1}%
\thanks{$^{*}$J. Nubert is also with ETH Zürich, Switzerland.}%
\thanks{$^{\ddagger}$H.\ Hose and S.\ Trimpe are with the Institute for Data Science in Mechanical Engineering (DSME), RWTH Aachen University, Germany.}%
}

\begin{document}

\maketitle
\thispagestyle{empty}
\pagestyle{empty}

\begin{abstract}
Approximating \ac{MPC} using \ac{IL} allows for fast control without solving expensive optimization problems online.
However, methods that use neural networks in a simple L2-regression setup fail to approximate multi-modal (set-valued) solution distributions caused by local optima found by the numerical solver or non-convex constraints, such as obstacles, significantly limiting the applicability of approximate MPC in practice.
We solve this issue by using diffusion models to accurately represent the complete solution distribution (i.e., all modes) up to kilohertz sampling rates.
This work shows that \acl{DAMPC} significantly outperforms L2-regression-based \acl{AMPC} for multi-modal action distributions. In contrast to most earlier work on \acs{IL}, we also focus on running the diffusion-based controller at a higher rate and in joint space instead of end-effector space.
Additionally, we propose the use of gradient guidance during the denoising process to consistently pick the same mode in closed loop to prevent switching between solutions.
We propose using the cost and constraint satisfaction of the original \acs{MPC} problem during parallel sampling of solutions from the diffusion model to pick a better mode online.
We evaluate our method on the fast and accurate control of a 7-DoF robot manipulator both in simulation and on hardware deployed at 250 Hz, achieving a speedup of more than 70 times compared to solving the MPC problem online and also outperforming the numerical optimization (used for training) in success ratio.\looseness=-1
\end{abstract}


\section{Introduction}
\input{Chapters/1_Introduction}

\section{Related Work}

\input{Chapters/2_Related_Work}

\section{Problem Formulation}
\input{Chapters/3_Problem_Formulation}

\section{Method: Diffusion-Based Approximate MPC}
\input{Chapters/4_Method}

\section{Setup \& Implementation Details}
\input{Chapters/5_Implementation}

\section{Results}
\label{sec:results}
\input{Chapters/6_Results}

\section{Conclusions \& Future Work}
\input{Chapters/7_Conclusions}



\section*{Acknowledgments}

We thank F.\ Grimminger, V.\ Berenz, B.\ Javot, and O.\ B.\ Aladag for supporting our experiments with the Apollo robot.

\bibliographystyle{IEEEtran}
\bibliography{references}

\end{document}

%% file: custom/packages.tex
\usepackage[dvipsnames,svgnames]{xcolor}
\usepackage{graphicx} 
\usepackage{amsmath} 
\usepackage{amssymb}  
\usepackage{mathtools}
\usepackage[inkscapelatex=true]{svg}
\usepackage[utf8]{inputenc}
\usepackage[OT1]{fontenc}
\usepackage{colortbl}
\usepackage[nospace]{cite}
\usepackage{booktabs}
\usepackage{lipsum}  
\usepackage{pdfpages}
\usepackage{makecell}
\usepackage{acronym}
\usepackage{siunitx}
\usepackage{multirow}
\usepackage{tabularx}
\makeatletter
\let\NAT@parse\undefined
\makeatother
\usepackage[colorlinks=true, citecolor=black, linkcolor=black, urlcolor=black]{hyperref}
\usepackage{cleveref}
\usepackage{wasysym}

\usepackage[absolute,overlay]{textpos} 

%% file: custom/acronyms.tex
\acrodef{MPC}{model predictive control}
\acrodef{AMPC}{approximate MPC}
\acrodef{DAMPC}{diffusion-based AMPC}
\acrodef{TO}{trajectory optimization}
\acrodef{IL}{imitation learning}
\acrodef{DM}{diffusion model}
\acrodef{DDPM}{Denoising Diffusion Probabilistic Model}
\acrodef{DDIM}{Denoising Diffusion Implicit Model}
\acrodef{LSM}{least squares model}
\acrodef{ATE}{absolute translation error}
\acrodef{ARE}{absolute rotation error}
\acrodef{TRT}{transient response time}
\acrodef{SR}{success rate}
\acrodef{SNR}{signal-to-noise ratio}
\acrodef{DOF}{degree-of-freedom}
\acrodef{LS}{least squares}
\acrodef{MLP}{multilayer perceptron}
\acrodef{ES}{early stopping}
\acrodef{GG}[$\nabla$G]{gradient guidance}

%% file: custom/commands.tex
\definecolor{gray}{gray}{0.3}
\colorlet{CaptionColor}{gray!15}
\DeclareMathOperator*{\argmin}{arg\,min}

\crefname{paragraph}{paragraph}{paragraphs}
\Crefname{paragraph}{Sec.}{Secs.}
\Crefname{chapter}{Chap.}{Chaps.}
\Crefname{section}{Sec.}{Secs.}
\Crefname{subsection}{Sec.}{Secs.}
\Crefname{figure}{Fig.}{Figs.}
\Crefname{table}{Tab.}{Tabs.}
\Crefname{equation}{Eq.}{Eqs.}
\Crefname{appendix}{App.}{Apps.}

\newcommand{\SEthree}{\mathrm{SE}(3)}

%% file: Chapters/1_Introduction.tex
Fast and dynamic systems are ubiquitous in robotics, ranging from autonomous vehicles to agile manipulation, where high-frequency control is essential for stability and precision.
While diffusion models excel at high-level planning~\cite{diffusion_policy,planning_diffusion,carvalho2023motion}, low-level dynamic control tasks have not yet seen comparable advancements.
This gap is particularly critical for systems requiring computationally efficient controllers that provide theoretical guarantees.\looseness=-1

\ac{MPC} is a versatile approach to control constrained linear and nonlinear systems~\cite{mpc}.
By formulating a constrained optimization problem that is solved iteratively in closed loop, \ac{MPC} provides stability guarantees, constraint satisfaction, and thus safety, and it has been successfully applied to a broad class of control problems.
However, solving \ac{MPC} optimization problems online remains a computational challenge in applications with fast dynamics, high update rates, complex formulations like nonlinear or robust \ac{MPC}, and deployment on resource-constrained embedded platforms.
Several teams have bridged this gap with \ac{AMPC}, which bypasses online optimization by learning explicit controller representations through \ac{IL}~\cite{Gonzalez2023AMPC_review}.
A common approach is to fit a \ac{MLP} with \ac{LS} regression~\cite{hose, PARISINI1995regression_ampc, LUCIA2018regression_ampc, BONZANINI2021regression_ampc, drgona2022AMPC_mpcobjective, nubert}, which has shown promising results in practical settings~\cite{nubert,abu2022deep,hose2024parameter,hose2025miniwheelbot}.
\pdfpxdimen=\dimexpr 1 in/72\relax
\begin{figure}[t]
\centering
\includegraphics[width=\columnwidth]{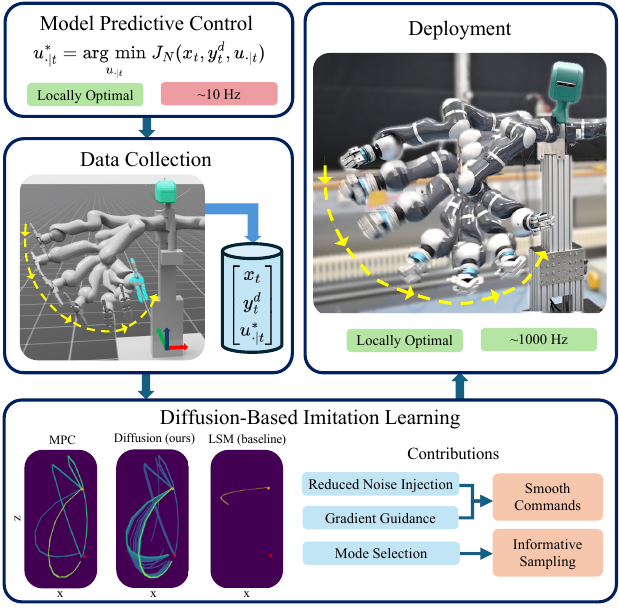}
\caption{
The \acl{DM} uses IL to approximate the multi-modal action distribution of \ac{MPC} optimization problems (lower left).
We highlight advantages over classical \ac{AMPC} and achieve high-quality solutions at fast update rates by \textit{i)} stopping noise injection early, \textit{ii)} using gradient guidance for closed-loop consistency, and \textit{iii)} selecting the best mode online.\looseness=-1
}
\label{fig:fig1}
\vspace{-3ex}
\end{figure}
Yet, one of the main issues experienced with \ac{AMPC} based on a \ac{LSM} is multi-modality in the mapping from state to expert trajectories~\cite{li2022using}.
When the \ac{MPC} is formulated as a nonlinear and non-convex optimization problem, the solution is not guaranteed to be unique, as multiple or infinitely many local and/or global solutions can exist.
Further, typical numerical solvers find only locally optimal solutions, which may lead to multi-modal (set-valued) distributions of a given state's \ac{MPC} control commands.
Fundamentally, \ac{LS} models cannot learn set-valued solutions~\cite{bishop1994mixture} provided by the MPC expert and instead learn the non-optimal, and not necessarily feasible, mean of the conditioned target distribution.\looseness=-1

In this work, we propose the use of diffusion models to approximate multi-modal solution distributions in \ac{AMPC} (\Cref{fig:fig1}).
We can gracefully approximate inherently set-valued problems and local optima from numerical solvers.
However, as shown in \Cref{sec:results}, naively sampling from the approximated solution distribution of the diffusion model leads to jerky and inconsistent robot behavior, as the policy output can alternate between different modes.
We propose using \ac{GG} and \ac{ES} of noise injection \textit{during} denoising to achieve consistent and smooth action trajectories in closed loop.
Further, \textit{after} denoising we propose to \textbf{\textit{i)}} check cost and constraint satisfaction of proposed solutions or \textbf{\textit{ii)}} select the correct mode from the diffusion model through democratic voting.
To evaluate our approach, we approximate a nonlinear \ac{MPC} formulation for $\SEthree$ setpoint tracking of a 7-\ac{DOF} robotic manipulator in simulation and the real world including obstacle avoidance, commanding joint velocities from a given initial joint configuration. 
This task involves multiple local optima and set-valued global optima due to the non-convexity of the \ac{MPC} and the robotic arm's additional \ac{DOF}.
The proposed \ac{DAMPC} significantly outperforms standard regression models in tasks with multiple solution modes. Interestingly, the \ac{DM} can even outperform the teacher \ac{MPC} on some metrics.\looseness=-1

As summarized in the video attached to this paper and the project website\footnote{Project website with code: \href{https://paumarquez.github.io/diffusion-ampc}{https://paumarquez.github.io/diffusion-ampc}}, the contributions of this work are:
\begin{enumerate}
    \item A novel diffusion-based \ac{AMPC} formulation that captures multi-modal action distributions and significantly outperforms L2 regression.\looseness=-1
    \item A closed-loop \ac{AMPC} framework 
    capable of running at up to \SI{1}{\kilo\hertz}, including \textit{\textbf{i)}} a denoising strategy to enhance closed-loop performance by improving consistency and smoothness (\ac{GG} and \ac{ES}) and \textit{\textbf{ii)}} parallel distribution sampling to either pick the best sample or perform democratic voting.\looseness=-1
    \item First demonstration of diffusion-based \ac{AMPC} on a real-world robotic system and comparison to \ac{LSM}.\looseness=-1
\end{enumerate}


%% file: Chapters/2_Related_Work.tex
\paragraph*{\ac{AMPC}}
Previous efforts in \ac{AMPC} have focused on \ac{IL} by collecting datasets of locally optimal \ac{MPC} solutions and training a machine-learning model that predicts \ac{MPC} control inputs.
Many focus on machine-learning models trained with classic (\textit{L2}-)regression~\cite{nubert,hose,Luci2021regression_ampc,PARISINI1995regression_ampc, Tulsyan2018Regression_ampc}; also see~\cite{Gonzalez2023AMPC_review} for a survey.
While these models work for unimodal distributions, they cannot approximate the multi-modal distributions commonly found in non-convex \ac{MPC}.
Thus, multi-modal solutions must be avoided altogether in the dataset~\cite{li2022using}. 
Multi-modal distributions in \ac{AMPC} have already been approached using mixture density networks~\cite{okamoto2024deep, Carius2019MPCNetAF}.
While these models are fast to evaluate, they require the number of modes as a hyperparameter of the model, which is difficult to know in practice.
In contrast, diffusion can approximate arbitrary distributions in a single model.\looseness=-1

\paragraph*{\ac{DM}}
Previous investigations of diffusion models for robotics have primarily focused on higher-level planning by mapping RGB-D sensor data to expert actions from human demonstrations.
Diffusers~\cite{planning_diffusion} use diffusion models to learn the dynamic model of a robotic arm and employ guidance~\cite{Dhariwal2021Guidance} to describe the task at hand by steering the sampling distribution to sample trajectories with desired initial and final states. Yet, \cite{Dhariwal2021Guidance} proposed to learn the joint distribution of state and action pairs, forcing the system to infer future states for every denoising step, adding computational overhead.
In contrast, diffusion policy~\cite{diffusion_policy} is a diffusion model that learns the conditional distribution of the actions given the current observation; these authors performed real-time high-level end-effector position and velocity planning at \SI{10}{\hertz} in a receding horizon control fashion.
Motion planning diffusion~\cite{carvalho2023motion} learns a prior model over trajectories to perform robot navigation and injects gradient guidance for obstacle avoidance.
Consistency policy~\cite{prasad2024consistencypolicy} used consistency models to distill a diffusion policy to use fewer steps (between $1$ and $3$ denoising steps), allowing a much faster control rate.
In this work, we propose a model that learns from \ac{MPC} demonstrations while directly commanding joint-level actions. We use gradient guidance to achieve smooth trajectories without mode swaps.\looseness=-1

\paragraph*{Diffusion-based \ac{MPC}}
Recent works have used diffusion models to approximate an \ac{MPC} formulation.
Similar to our method, \cite{Huang2024DiffusionAMPC} and \cite{Romer2024DiffusionAMPCConstraints} propose imitation-learning approaches.
The former finds globally optimal solutions similar to multi-start nonlinear solvers.
However, the computational time of the approximation (more than \SI{200}{\milli\second}) is too large to control general dynamic robotic systems.
The latter uses a diffusion model to learn trajectories of state-action pairs similarly to \cite{planning_diffusion}, and it generates safe trajectories with new constraints not present in the training dataset.
Finally, \cite{Li2024DiffuSolve} trains a \ac{DM} using imitation learning and uses the approximation to warm-start an optimization solver.
While they improve computational time, it remains unbounded with a high standard deviation, preventing this method from being deployed reliably in critical real-time systems.
To our knowledge, we are the first to apply diffusion-based \ac{AMPC} on real robotic hardware for low-level control at high update rates.\looseness=-1

%% file: Chapters/3_Problem_Formulation.tex
We consider general, nonlinear, discrete-time systems $x_{t+1}=f(x_{t}, u_{t})$,
with state $x_t \in \mathcal{X} \subseteq \mathbb{R}^n$, input $u_t\in \mathcal{U} \subseteq \mathbb{R}^m$, time index $t \in \mathbb{N}$, and dynamics function $f: \mathcal{X} \times \mathcal{U} \mapsto \mathcal{X}$.
The control objective is to drive the system to a reference point where $h(x_t)=y_t^d$ and $f(x_t, u_t)=x_t$, where $y_t^d \in \mathcal{Y} \subseteq \mathbb{R}^n_y$ is the desired output and $h: \mathcal{X} \mapsto \mathcal{Y}$ the output function.\looseness=-1

\subsection{Model Predictive Control in Simulation}
We assume the existence of a general nonlinear \ac{MPC} or \ac{TO} formulation that can solve the specified task under perfect assumptions in simulation:
\begin{align}
    \begin{split}
    \label{eq:opt_problem}
    u_{\cdot|t}^* := &\argmin_{u_{\cdot|t},x_{\cdot|t}} J_N(u_{\cdot|t}; x_t, y_t^d) \\
    \text{s.t.~} & x_{0|t} = x_t,  \\
    & x_{k+1|t} = f(x_{k|t}, u_{k|t}), \\
    & g_j(x_{k|t}, u_{k|t}) \leq 0. 
    \end{split}
\end{align}
Here, $N$ is the prediction horizon with $k\in \left[0, 1, ..., N-1\right] \in \mathbb{N}$, state $x_{\cdot|t} \in \mathcal{X}^N$, input $u_{\cdot|t} \in \mathcal{U}^N$, cost function $J_N: \mathcal{U}^N \mapsto \mathbb{R}$, and constraints $g_j: \mathcal{X} \times \mathcal{U} \mapsto \mathbb{R}$ with $p$ constraints.
The feasible set $\mathcal{Z}$ is defined as\looseness=-1
\begin{equation}
    \mathcal{Z} = \{(x, u) \in \mathcal{X} \times \mathcal{U} \;|\; g_j(x, u) \leq 0, j = 1, 2, ..., p\}.
\end{equation}
In simulation, an \ac{MPC} controller solves \Cref{eq:opt_problem} for each $t$ and applies the policy $\pi_{\text{MPC}}(x_t) \coloneqq u_{0|t}^*$ in closed loop.\looseness=-1

\subsection{Local Optimal Solver}
We assume the availability of a local optimal solver of \Cref{eq:opt_problem}, denoted as $s$.
In practice, $s$ can be based on sequential quadratic programming, interior point methods, or even sampling-based zero-order methods.
The corresponding solution depends on the initialization $\xi^\text{init}$ of the optimization variables $\xi \in \Xi$ and parameters $x_t$ and $y^d$.
We do \textit{not} assume to have an informed initial guess.
The set of distributions of locally optimal solutions $u^*_{\cdot|t} \coloneqq s(x_t, y^d, \xi^\text{init}) \in \mathcal{U}^N$ found by the solver $s$ is given as\looseness=-1
\begin{equation}
\label{equ:distribution_tb_learned}
    Q_s = \Bigl\{s(\xi^\text{init}, x_t, y^d) \;\Big|\; \xi^\text{init} \sim \text{P}(\Xi), x_t\in\mathcal{X}, y^d_t\in\mathcal{Y}\Bigr\},
\end{equation}
where $\text{P}(\Xi)$ denotes an arbitrary distribution over the optimization parameters $\Xi$.

\subsection{Objective}
We can sample from the set $Q_s$ conditioned on $x_t$ and $y^d$ as $u^*_{\cdot|t} \sim Q_s(\cdot | x_t, y^d_t)$. Note that this policy is assumed to be non-deterministic, in contrast to most other \ac{AMPC} formulations, e.g.,~\cite{nubert}.
This work aims to find an explicit approximation to reliably sample from $Q_s$ at a fast update rate without numerically solving~\Cref{eq:opt_problem}.\looseness=-1 

%% file: Chapters/4_Method.tex
This section describes the proposed \ac{DAMPC} formulation of using a \ac{DM} to sample from \Cref{equ:distribution_tb_learned}.
The full system includes \textit{data collection} (\Cref{sec:method_data_collection}), \ac{DM} prior \textit{training} (\Cref{sec:method_diffusion}), and \ac{DM} \textit{sampling} during inference (\Cref{sec:method_inference}).\looseness-1

\subsection{Data collection}
\label{sec:method_data_collection}

We collect a dataset $\mathcal{D}$ consisting of $N_e$ episodes composed of the current state, desired setpoint, and locally optimal commands for $N_d$ steps per episode. $\mathcal{D}$ is given as $\mathcal{D} = \bigcup_{k=1}^{N_e} \mathcal{D}_k$ with each episode defined as
\begin{equation}
\begin{aligned}
\mathcal{D}_k &= \Bigl\{ \bigl( {}_k x_t,\; y^d_t,\; {}_k u_{\cdot\mid t}^* \bigr) \;\Big|\; 
                {}_k x_t \in \mathcal{X},\; {}_k y^d_t \in \mathcal{Y}, \\
                &\qquad u_{\cdot\mid t}^* \sim Q_s(\cdot \mid {}_k x_t,\; {}_k y^d_t),\; 
                t = 1, \dots, N_d \Bigr\}.
\end{aligned}
\end{equation}
While storing optimal actions~$u^*$, the generation of samples is performed by adding noise to the optimal \ac{MPC} policy to reduce distribution shift \cite{Spencer2021distribution_shift} according to\looseness=-1
\begin{align}
    \pi_{\text{MPC}}'(x_t) &:= \pi_{\text{MPC}}(x_t) + \sigma_{u_t}\epsilon, \quad \epsilon \sim \mathcal{N}(\mathbf{0}, \mathbf{I}),
\end{align}
with~$\sigma_{u_t} := \max\left(\tfrac{|\pi_{\text{MPC}}(x_t)|}{\text{SNR}^d}, \sigma_{\text{min}}\right)$ where SNR$^d$ is the desired signal-to-noise ratio. Here, $\sigma_{\min}$ is the minimum standard deviation of the noise added to explore regions of attraction near steady states.
The policy $\pi_{\text{MPC}}'$ is executed in simulation for $N_d$ step episodes, $N_e$ times.
Unsatisfactory local minima are removed from the training dataset.


\subsection{Diffusion-based \acs{MPC} Prior Model}
\label{sec:method_diffusion}
We train a diffusion model $f_\theta^{x_0}$ on the collected dataset $\mathcal{D}$ to draw samples from the distribution $u_{\cdot\mid t}^* \sim Q_s(\cdot \mid x_t, y^d_t)$.
We model the denoising process using Langevin dynamics and follow the standard \ac{DDPM}~\cite{Ho2020DDPM}.
Here, a model $f_\theta^{x_0}(x_i, i)$ is trained to predict the clean sample $x_0$ from a given noisified sample $x_i$ after $i$ noisy steps in the \textit{forward process}\looseness=-1
\begin{equation}
    q(x_{i}|x_{i-1}) := \mathcal{N}(x_i; \sqrt{1-\beta_i}x_{i-1}, \beta_i\mathbf{I}),
\end{equation}
where $\beta_i$ is a parameter that adjusts the \ac{SNR} at each step $i$ depending on the chosen noise schedule.
The training loss for $f_\theta^{x_0}$ can be derived and simplified from the variational bound on negative log likelihood (see~\cite{Ho2020DDPM} for details):
\begin{equation}
    \mathcal{L}_{\text{ELBO}}^{x_0} := \|x_0 - f_\theta^{x_0}(\sqrt{\bar{\alpha}_i}x_0 + \sqrt{1-\bar{\alpha_i}}\epsilon_i, i)\|^2.
\end{equation}



\subsection{Diffusion Model Sampling for Inference}
\label{sec:method_inference}

Once $f_\theta^{x_0}$ is trained, samples from the underlying distribution of the training dataset can be drawn by gradually denoising an initial $x_{N_I} \in \mathcal{N}(\mathbf{0}, \mathbf{I})$ following
\begin{align}\label{eq:denoising_x0}
    x_{i-1} &:= \tfrac{\sqrt{\bar{\alpha}_{i-1}}\beta_i}{1 - \bar{\alpha}_i}f_\theta^{x_0}(x_i, i) + \tfrac{\sqrt{\alpha_{i}}(1-\bar{\alpha}_{i-1})}{1 - \bar{\alpha}_i}x_i + \tilde{\beta}_i \epsilon,
\end{align}
where $\tilde{\beta}_i := \tfrac{1 - \bar{\alpha}_{i-1}}{1 - \bar{\alpha}_i}\beta_i$ and $\epsilon \sim \mathcal{N}(\mathbf{0}, \mathbf{I})$.
This formulation is mathematically equivalent to predicting the added noise $\epsilon_i$, instead of the clean sample $x_0$.
The mapping from predicted noise to clean sample is given by
\begin{equation}
    x_0=(x_i - \sqrt{1-\bar{\alpha}_i}\epsilon)/\sqrt{\bar{\alpha}_i},
    \label{eq:diff_x_epsilon}
\end{equation}
which is relevant to exploit the diffusion model's relationship with score matching to allow for using \ac{GG}~\cite{Vincent2011DenoisingandScorematching}:
\begin{equation}
    \nabla_x \log p(x) = - \frac{\epsilon}{\sqrt{1-\bar{\alpha}_t}}.
    \label{eq:diff_score}
\end{equation}
In the following, we suggest three improvements on the denoising process during inference over \ac{DDPM}~\cite{Ho2020DDPM}.

\subsubsection{Consistent Mode Selection through Gradient Guidance}
\label{sec:method_guidance}
In optimization-based real-time \ac{MPC}, numerical solvers often show consistent mode selection in closed loop due to warm-starting.
However, sampling from a diffusion model approximating multiple local solutions can lead to random mode selection in subsequent steps, causing inconsistent and jerky control commands.
Inspired by warm-starting in real-time optimization, we propose to use \textit{\acl{GG}} for the \ac{DM} to condition the distribution of the control commands to the previously executed command $p(u_t^*\mid x_t, y^d,u_{t-1}^*)$.
Given a diffusion model that samples from $Q_s$, we transform $f_\theta^{x_0}$ into $f_\theta^{\epsilon}$ by using \Cref{eq:diff_x_epsilon} to exploit its relationship with score matching (\Cref{eq:diff_score}).
Given $p(x_{i-1\mid t}\mid x_{i\mid t}, i)$ modeled by the \ac{DM}~\cite{Ho2020DDPM}, where $i$ refers to the step during the denoising process, we can add prior information $o$ by using Bayes' Rule~\cite{Dhariwal2021Guidance}:\looseness=-1
\begin{equation}
    p(x_{i-1\mid t}|x_{i\mid t}, i, o) \propto p(x_{i-1\mid t}|x_{i\mid t}, i)p(o|x_{i \mid t}, i).
\end{equation}
By using the log's property, this can be written as
\begin{equation}
\begin{aligned}
    \nabla_x \log p(x_{i-1 \mid t}|x_{i\mid t}, i, o) \propto &\nabla_x \log p(x_{i-1 \mid t}|x_{i\mid t}, i) \\
    &+ \nabla_x \log p(o|x_{i\mid t}, i).
    \label{eq:log_guidance_bayes}
\end{aligned}
\end{equation}
Therefore, we can guide the distribution toward the previous mode to ensure closed-loop consistency by redefining $f_\theta^\epsilon$ as
\begin{equation}
\begin{aligned}
    f^\epsilon_\theta(x_{i\mid t}, i, o) :=& f^\epsilon_\theta(x_{i\mid t}, i) - \\
    &\nabla_{x_{i\mid t}} \log \mathcal{N}\left(x_{i\mid t};x_{0\mid t-1}, (1-\bar{\alpha}_i)\mathbf{I} \right).
\end{aligned}
\end{equation}

\subsubsection{Reduced Jerk by Stopping Noise Injection Early}
\label{sec:method_jerk}
During denoising, randomness is added via the noise schedule~$\tilde{\beta}_i$ in \Cref{eq:denoising_x0}, balancing smoothness and sample diversity.
For high-frequency control, smooth solutions are prioritized over exploration to prevent vibrations. Thus, we propose to perform denoising as\looseness=-1
\begin{align}
\label{equ:early_stopping}
    x_{i-1} &:= \frac{\sqrt{\bar{\alpha}_{i-1}}\beta_i}{1 - \bar{\alpha}_i}x_0 + \frac{\sqrt{\alpha_{i}}(1-\bar{\alpha}_{i-1})}{1 - \bar{\alpha}_i}x_i + \tilde{\beta}_i' \epsilon,
\end{align}
where $\tilde{\beta}_i' := \tilde{\beta}_i \mathbf{1}_{i > i_\text{min}^\epsilon}$ and $\epsilon \sim \mathcal{N}(\mathbf{0}, \mathbf{I})$.
Adding noise at the lower \ac{SNR} steps allows the diffusion model to explore multi-modality. 
Not adding the noise at the higher \ac{SNR} steps allows the closed-loop function to be smooth, considerably reducing the resulting trajectory's jerk.


\subsubsection{Constraint Satisfaction via Sampling in Closed Loop}
\label{sec:method_sampling}
We propose methods for informative sampling from $Q_s(\cdot \mid x_t, y^d_t)$ to reduce approximation uncertainty and improve sample fidelity.
Similar to multi-start methods~\cite{Huang2024DiffusionAMPC} for nonlinear optimization, we generate $L$ inputs $\{u_l^* \mid u_l^* \sim Q_s(\cdot \mid x, y^d)\}_{l=1}^L$ by sampling the diffusion model (leveraging GPU parallelization) and suggest \textit{two} ways to rank these samples.\looseness=-1

\paragraph{Full State Knowledge Available}
We can score and rank the samples according to feasibility and (\ac{MPC}) cost.

\paragraph{No Full State Knowledge}
When we cannot reliably estimate the cost of each sample, e.g., when the diffusion model is implicitly estimating the state of the system, we propose clustering the samples based on a similarity metric, such as the Euclidean distance in the command space, or a domain-specific distance measure.
Let $\mathcal{C} = \{C_1, C_2, ..., C_K\}$ represent $K$ clusters, where each cluster $C_k$ corresponds to a distinct mode of the distribution.
The probability mass of each cluster is approximated by the empirical density $P(C_k) \approx \frac{|C_k|}{L}$.
The selected trajectory $u^*_l$ is drawn from the highest-density cluster, ensuring that the chosen trajectory is representative of the most probable mode in the distribution, discarding low-density samples that commonly correspond to unsuccessfully denoised samples or unlikely local minima generated by the \ac{MPC} during data collection.\looseness=-1

%% file: Chapters/5_Implementation.tex
To assess the performance of our method, we approximate an \ac{MPC} formulation for full $\SEthree$ end-effector tracking of a 7-\ac{DOF} \textit{KUKA LBR4+} arm (\Cref{fig:fig1}) by commanding joint velocities at a high rate.
This task exposes multi-modality due to the redundant \ac{DOF}, obstacles in the environment, and the implicit non-convexity of the optimization problem.
In this section, we explain the \ac{MPC} formulation and the design parameters of the data collection, baselines, and \ac{DAMPC}.\looseness=-1

\subsection{Model Predictive Control for End-Effector Pose Tracking}
\label{sec:setup_mpc}
We use an \ac{MPC} formulation based on~\cite{nubert,LIMON2018MPCSetpoint,KOHLER2020Dynamic} extended to track full $\SEthree$ end-effector poses.
We define a state $x_t \in \mathcal{X} \subset \mathbb{R}^7$ as the joint positions and $u_t \in \mathcal{U} \subset \mathbb{R}^7$ as the joint velocities.
As opposed to \cite{nubert}, and following \cite[Ass.~1]{KOHLER2020Dynamic}, we define a virtual trajectory $x^s_{\cdot|t} \in \mathcal{X}^{N+1}$ and input $u^s_{\cdot|t} \in \mathcal{U}^{N+1}$, with a reference $y_t \in \SEthree = \mathcal{Y}$ and virtual output $y^s=h(x^s_{N|t})$.
The control objective is the minimization of the tracking error $e_t=x^s_{N|t}-x_t$ and constraint satisfaction $(x_t, u_t) \in \mathcal{Z}$.
We found that a virtual trajectory instead of a virtual terminal state as in~\cite{nubert} improved convergence. The corresponding MPC formulation for output tracking is:
\begin{align}
\begin{split}
   \min&_{u_{\cdot|t}, u^s_{\cdot|t}, x^s_{\cdot|t}} J_N(u_{\cdot|t}, u^s_{\cdot|t}, x^s_{\cdot|t}; x_t, y_t^d) \\
   \label{eq:opt_virutal_problem}
   \text{s.t.~} & x_{0|t} = x_t, \quad
   x^s_{0|t} = x_t \\
   & x_{k+1|t} = f(x_{k|t}, u_{k|t}), \quad
   x^s_{k+1|t} = f(x^s_{k|t}, u^s_{k|t}), \\
   & g_j(x_{k|t}, u_{k|t}) \leq 0, \quad
   g_j(x^s_{k|t}, u^s_{k|t}) \leq 0, \\
   & f(x^s_{N|t}, u^s_{N|t}) = x^s_{N|t},
\end{split}
\end{align}
for $k= 0, 1, ..., N-1$, where the cost is
\begin{align}
\begin{split}
   J_N(u_{\cdot|t}, u^s_{\cdot|t}; x_t, y_t^d) 
   := &
   \left\lVert x_{N|t} - x^s_{N|t}\right\rVert_\mathbf{P} 
    + \\ 
    d_y(y_t^d, h(x^s_{N|t}))  + & \sum_{k=0}^{N-1} l(x_{k|t}, u_{k|t}, x^s_{N|t}, u^s_{N|t}),
\end{split}
\end{align}
with quadratic stage cost $l(x, u, x^d, u^d)= \left \lVert x - x^d \right\rVert_\mathbf{Q}^2 + \left \lVert u - u^d \right\rVert_\mathbf{R}^2$ with positive definite $\mathbf{Q}$ and $\mathbf{R}$.
The dynamics function $f$ is the Euler integration of the joint position $x_{k|t}$ with the joint velocity commands $u_{k|t}$.
The output error $d_y: \mathrm{SE}(3) \times \mathrm{SE}(3) \mapsto \mathbb{R}$ is given by a weighted average of the position distance $\|y_p - h(x^s)_p\|^2_2$ and the rotation distance $\|\ (\log(y_R^{-1} h(x^s)_R))_\vee\|^2_2$, where the subindex $p$ and $R$ denote the position and rotation components of $\SEthree$ elements.
Furthermore, $\log(\cdot)$ refers to the mapping from the Lie group $\mathrm{SE}(3)$ to the Lie algebra $\mathfrak{se}(3)$, and $[\cdot]_\vee$ is the projection from the Lie algebra to the tangent space.
The constraint set $\mathcal{Z}$ is defined with polytopic constraints on the state and command limits, and spherical obstacle avoidance constraints\looseness=-1
\begin{equation}
    \|(h(x_{k|t}^i)_p - o^p_j) \odot o^s_j\|_2 \geq 1
\end{equation}
on the output space $\forall$ time steps $k \in [0, .., N-1] \subset \mathbb{N}$, joints $i \in [1, .., 7] \subset \mathbb{N}$, and obstacles $j \in [1, ..., M] \subset \mathbb{N}$ with $M$ obstacles.
The ellipsoidal regions representing the robot's body and external objects are defined by position $o^p_j \in \mathbb{R}^3$ and scaling $o^s_j \in \mathbb{R}^3$, with $\odot$ denoting element-wise product.

\subsection{Data Collection}
We collect two datasets: the first has only self-collision avoidance, and the second adds a spherical obstacle in the center of the task space to assess non-convex constraint satisfaction of the approximation.
We collect the datasets with the method explained in \Cref{sec:method_data_collection} for $N_d=80$ step episodes with random initial joint positions $x_0 \in \text{Unif}(\mathcal{X})$ and reachable targets $y^d \in \text{Unif}(\mathcal{Y}^d)$ where $\mathcal{Y}^d \coloneqq \{h(x) \mid x \in \mathcal{X}\}$.
The final datasets contain $55.5$ and $33.9$ million open-loop \ac{MPC} predictions, respectively.
The prediction horizon is $N=20$ and $\Delta t=\SI{0.1}{\second}$.
The state constraints are given by the joint limits, and the input constraints by the \textit{KUKA}'s safety stop, which engages beyond $2.3\tfrac{\text{rad}}{\text{s}}$.
The noise coefficients for exploration are \ac{SNR}$^d=0.8$ and $\sigma_{\min}=0.35\tfrac{\text{rad}}{\text{s}}$.
These values were chosen empirically by increasing noise to the maximum level that allows the target to be reached.
We use IPOPT~\cite{Wachter2006ipopt} as solver $s$ and uniformly randomize the initial guess $\xi_\text{guess}$ at $t=0$. 
For subsequent steps, we warm-start the solver with the previous solution to ensure mode diversity in the dataset.
We excluded unsatisfactory solutions where $d_y(y_t^d, h(x^s_{N|t})) > 0.01$, which we attribute to the random initial guess, thereby removing 5\% of the dataset.\looseness=-1


\subsection{Implementation Details}
The data collection pipeline is built on top of Isaac Lab \cite{orbit} with expert solutions computed with IPOPT~\cite{Wachter2006ipopt} on CPU using CasADi~\cite{Andersson2019CasADi} and Pinocchio~\cite{Carpentier2019Pinocchio}.
Using multi-processing techniques, we deploy CoClusterBridge
to parallelize the data collection on the vectorized simulator.
For each dataset we train a \ac{DM} and a nonlinear \ac{LSM} as a baseline.
The observations of the models are the current joint state $x_t$ and the position and orientation of the target $y^d$, where the orientation is encoded with the $6$D representation provided by \cite{Zhou_2019_6drotation}.
The diffusion model is trained based on \ac{DDPM}~\cite{Ho2020DDPM} using a cosine noise scheduler \cite{nichol2021iDDPM} for $5$ steps with the rescaling proposed by~\cite{Lin2024flawed} to ensure $0$ \ac{SNR} at the last step.
We use the default $\gamma=5$ for the Min-\ac{SNR} weighting strategy~\cite{Hang2023MinSNR}.
For inference we use the proposed \ac{GG} (\Cref{sec:method_guidance}) to ensure consistency within subsequent closed-loop steps, and we set $i^\epsilon_\text{min}=\lfloor0.75 \cdot N_I \rfloor$ from \Cref{eq:denoising_x0} and \Cref{equ:early_stopping} to perform early stopping, where $N_I$ is the total number of denoising steps.
The model architecture is an \ac{MLP} composed of $7$ layers with $1,000$ neurons each, with a total of $6.6$ million trainable parameters (note that we did not optimize to reduce the size of this network), as well as \acp{MLP} for the temporal and observation encoder.
The baseline \ac{LS} model has $6$ layers and a total of $2.8$ million trainable parameters, tuned to minimize the validation loss.\looseness=-1

\subsection{Hardware Deployment}
We deploy the \ac{LSM} and the proposed \ac{DAMPC} controller with \ac{GG} and \ac{ES} on a \textit{KUKA LBR4+} robotic arm (Apollo, \cite{Kappler2018apollo}) at a control rate of \SI{250}{\hertz}.
We also compare to the optimization-based \ac{MPC}, which is limited to \SI{10}{\hertz} due to slow execution.
The policies run on an \textit{Intel Core i9-13900K} and an \textit{NVIDIA GeForce RTX 4070}; low-level PI controllers track the commanded joint velocities at \SI{1}{\kilo\hertz}.

%% file: Chapters/6_Results.tex
We evaluate the performance of the proposed diffusion-based approximation both in a kinematic simulation and on hardware experiments using Apollo. 
First, we evaluate the proposed denoising strategies of \Cref{sec:method_guidance} and \Cref{sec:method_jerk}, showing why the proposed variations of standard \ac{DDPM} are needed for hardware deployment.
Then, we analyze the informative sampling strategies from \Cref{sec:method_sampling} by leveraging the model's ability to provide a large set of samples.\looseness=-1

\subsection{Denoising Methods for Diffusion-based Control}
This section provides results for \ac{DDPM} with \ac{GG} and \ac{ES} (ours), as well as the following baselines: vanilla \ac{DDPM}~\cite{Ho2020DDPM}, \ac{DDIM}~\cite{ddim} to reduce denoising steps during inference, and \ac{LSM} (\ac{MLP} with L2-regression loss).
We quantitatively and/or qualitatively report each method's ability to represent multi-modal action distributions, tracking performance, the effectiveness of \ac{GG} and \ac{ES}, feasibility, and computational time via simulation and hardware experiments.\looseness=-1

\subsubsection{Representing Multi-Modality}
\label{sec:res:multimodal}
\begin{figure}[t]
\vspace{0.2cm}
\centering
\includegraphics[width=0.95\columnwidth]{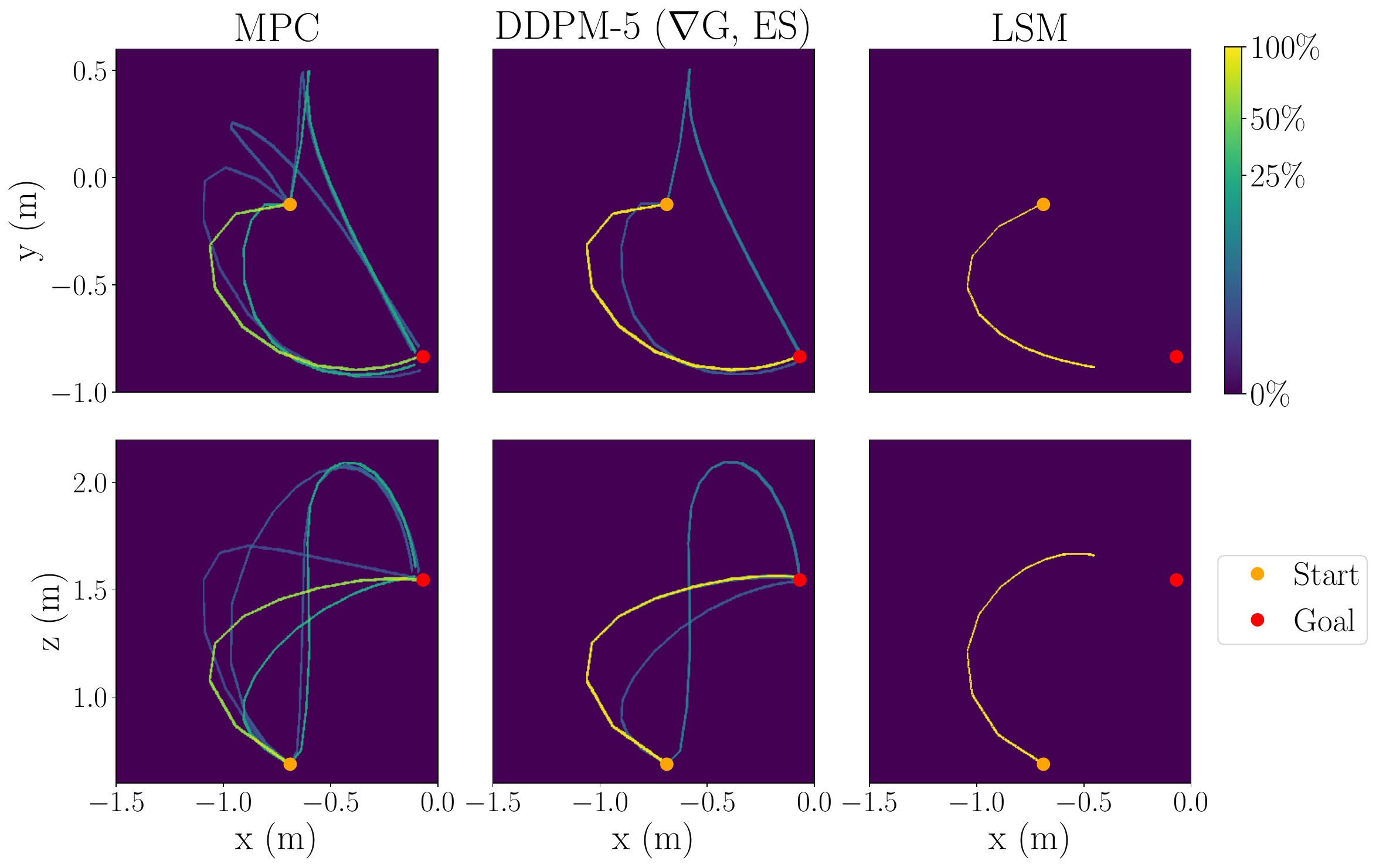}%
\vspace{-0.5em}
\caption{Heatmaps representing the end-effector position's probability density (log scale), projected onto the $x$-$y$ and $x$-$z$ planes over multiple runs for the same start $x_0$ and goal $y^d$. Qualitative multi-modality of (open-loop) \ac{MPC} solutions $x_{\cdot\mid 0}$ are shown on the left, which are approximated well by \ac{DDPM} (ours, center). In contrast, the \ac{LSM} (right) collapses to the dominant mode without reaching the target.
}
\label{fig:heatmap_lsm_mpc_clddpm}
\vspace{-3ex}
\end{figure}

We verify ap\-prox\-i\-ma-tion of the multi-modal distributions qualitatively in simulation.
As shown in \Cref{fig:fig1} and \Cref{fig:heatmap_lsm_mpc_clddpm} (left), the solution to the MPC optimization is a multi-modal distribution of open-loop end-effector trajectories (i.e., the predicted $x_{\cdot\mid 0}$ from \Cref{eq:opt_virutal_problem}) for different initial guesses~$\xi$ but the same initial state $x_0$ and desired output $y^d$.
Indeed, the \ac{LSM} cannot approximate multi-modal distributions, as shown in \Cref{fig:heatmap_lsm_mpc_clddpm} (right).
However, when one trajectory dominates the density, the \ac{LSM} can decently predict this primary mode.
The proposed \ac{DM} can capture multiple different modes and predict long-horizon trajectories (center of \Cref{fig:heatmap_lsm_mpc_clddpm}).
Notably, some modes of the MPC solution distribution are not reproduced by the diffusion model because we filter out local minima that do not reach the target (cf. \Cref{sec:setup_mpc}) and because the model has only $5$ denoising steps.
Models trained with more denoising steps, e.g., $40$, generally capture low-density modes more completely.\looseness=-1

\pdfpxdimen=\dimexpr 1 in/72\relax

\subsubsection{Tracking Performance}
\label{sec:res:trackingperformance}
We quantitatively evaluate tracking performance for 8,750 random start and goal poses in simulation and 100 in hardware experiments.
The \acp{SR}, defined as the proportion of trajectories that terminate with errors below \SI{20}{\milli\meter} and \SI{5.7}{\degree}, the \acfp{TRT}, \acfp{ATE}, and \acfp{ARE} thereof are reported in \Cref{tab:metrics_lsm_mpc_clddpm} for simulation and \Cref{tab:apollo_metrics_lsm_mpc_clddpm} for hardware experiments.
The \ac{SR} of the proposed method is higher than that of the optimization-based \ac{MPC} in both settings due to convergence issues that appear only occasionally.
Removing these from the training dataset allows our \ac{DM} to outperform the expert in \ac{SR}.
Regarding translation and orientation error at the end of successful runs, the \ac{MPC} achieves the best metrics in simulation and hardware, reaching \SI{0.5}{\milli\meter} error during the \SI{5}{\second} episode in the perfect simulations.
In contrast, the diffusion-based approximators achieve a steady-state error of about \SI{5}{\milli\meter}, likely due to the emphasis of the data collection process on dynamic regions and the added noise $\sigma_{\min}\gg0$.
The \ac{LSM} shows poor tracking quality and rarely converges as it fails to approximate multi-modal behavior.\looseness=-1

\begin{table}[t]
    \caption{Closed-loop tracking performance in \textbf{non-real-time} simulation for random start and goal positions. Mean \ac{ATE}, \ac{ARE} and \ac{TRT} are computed only for successful episodes, the rate of which we report under~\acs{SR}. The proposed approach is shown in \textcolor{blue}{blue}.}
    \setlength\tabcolsep{4.0pt}
    \label{tab:metrics_lsm_mpc_clddpm}
    \begin{center}
    \begin{tabularx}{\columnwidth}{l| >{\raggedright\arraybackslash}c >{\raggedright\arraybackslash}c >{\raggedright\arraybackslash}c| >{\centering\arraybackslash}X >{\raggedright\arraybackslash}c >{\raggedright\arraybackslash}c >{\raggedright\arraybackslash}c}
        \hline
        \rowcolor{CaptionColor} 
        Method & $N_I$ & $\nabla$G & ES & \ac{ATE} [\si{\milli\meter}] & \ac{ARE} [\si{\degree}] & \acs{SR} [\%] & \ac{TRT} [\si{\second}] \\
        \hline
        \multicolumn{4}{l|}{MPC} & \textbf{0.54}   & \textbf{0.02} & 93.18 & 1.18 \\ 
        \multicolumn{4}{l|}{LSM} & 13.84                                        & 3.70 & 10.26 & 2.54   \\
        \hline
        \textcolor{blue}{DDPM} & 5  & $\times$ & $\times$ & 5.63                  & 1.29 & 95.76 & \textbf{1.14}   \\
        DDPM & 5  & $\times$ & & 5.63                  & 1.29 & 95.71  & \textbf{1.14}   \\
        DDPM & 40 &          &        & 5.14                   & 1.21 & 96.18 & 1.18   \\
        DDPM & 5  &          &        & 5.47                  & 1.26 & 96.33  & 1.17   \\
        DDIM & 5  &          &        & 5.07                     & 1.19 & \textbf{96.41}  & 1.18   \\
        \hline
    \end{tabularx}
    \end{center}
    \label{tab:mpc_lsm_ddpm_metrics}
    \vspace{-3ex}
\end{table}

\begin{table}[t]
    \caption{Closed-loop \textbf{hardware} experiments for random start and goal positions. Mean \ac{ATE}, \ac{ARE} and \ac{TRT} are computed for successful episodes, which occur at the rate of \acs{SR}.
    DDPM without guidance is \textit{not deployable} due to mode swapping in subsequent closed-loop steps, which leads to high jerk and unsafe behavior.
    The proposed approach is shown in \textcolor{blue}{blue}.
    }
    \label{tab:apollo_metrics_lsm_mpc_clddpm}
    \setlength\tabcolsep{4.0pt}
    \begin{center}
    \begin{tabularx}{\columnwidth}{l| >{\raggedright\arraybackslash}c >{\raggedright\arraybackslash}c >{\raggedright\arraybackslash}c| >{\centering\arraybackslash}X >{\raggedright\arraybackslash}c >{\raggedright\arraybackslash}c >{\raggedright\arraybackslash}c}
        \hline
        \rowcolor{CaptionColor} 
        Method & $N_I$ & $\nabla$G & ES & \ac{ATE} [\si{\milli\meter}] & \ac{ARE} [\si{\degree}] & \acs{SR} [\%] & \ac{TRT} [\si{\second}] \\
        \hline
        \multicolumn{4}{l|}{MPC} & \textbf{3.71} & \textbf{0.40} & 87.80 & 1.71 \\
        \multicolumn{4}{l|}{LSM} & 15.77 & 3.94 & 12.90 & 2.62 \\
        \hline
        \textcolor{blue}{DDPM} & 5 & $\times$ & $\times$ & 6.16 & 1.29 & \textbf{93.00} & \textbf{1.34} \\
        \hline
    \end{tabularx}
    \vspace{-3ex}
    \end{center}
\end{table}

For the simulation results in \Cref{tab:metrics_lsm_mpc_clddpm}, we additionally report the performance of vanilla \ac{DDPM} with 40 and 5 denoising steps, and vanilla \ac{DDIM} with 5 denoising steps. 
While all of these show similar steady-state performance in simulation, we \textit{cannot} evaluate these controllers on hardware due to vibrations and jerky behavior caused by inconsistent mode selection (without \ac{GG}), as detailed in the following section.\looseness=-1

\begin{figure}[bt]
    \centering
    \vspace{-1.5ex}
    \includegraphics[width=\columnwidth, clip, trim=0cm 0.2cm 0cm 0.2cm]{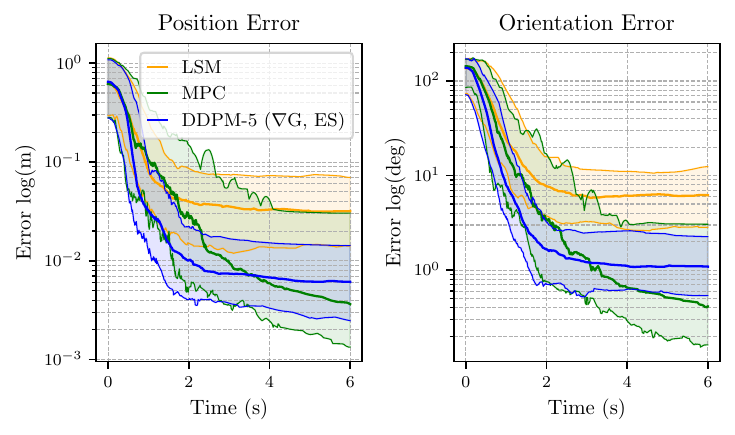}%
    \vspace{-1em}
    \caption{
    Tracking performance in closed-loop hardware experiments: mean end-effector position and orientation errors over time for the \ac{LSM}, the optimization-based \ac{MPC}, and the proposed diffusion model with five denoising steps, gradient guidance, and early stopped noise injection. The shaded areas show the 10th to 90th error percentiles.
    }
    \label{fig:apollo_time_error}
    \vspace{-3ex}
\end{figure}

We also report the evolution of the average tracking error over time for closed-loop experiments on hardware in \Cref{fig:apollo_time_error}.
The \ac{MPC} has poor transient performance and high overshoot, potentially leading to self-collisions, due to its slow update rate of \SI{10}{\hertz}.
In comparison, the transient performance of the proposed diffusion model and the \ac{LSM} benefit from the approximation's fast \SI{250}{\hertz} update rate; they show smooth behavior with little or no overshoot.
The \ac{LSM} 
has large overall tracking error due to multi-modality.

\begin{table}[bt]
    \caption{\ac{GG} and \ac{ES} noise injection in simulation. 
    Mode swaps are defined as the number of different local minima chosen.
    Statistics are extracted from episodes of \SI{2}{\second}.
    The median jerk norm shows the effectiveness of the noise schedule. The proposed method is shown in \textcolor{blue}{blue}.
    }
    \par\smallskip
    \centering
    \begin{tabularx}{\columnwidth}{l| >{\raggedright\arraybackslash}c >{\raggedright\arraybackslash}c >{\raggedright\arraybackslash}c| >{\centering\arraybackslash}X >{\raggedright\arraybackslash}c}
        \rowcolor{CaptionColor} 
        \hline
          Method & $N_I$ & \ac{GG} & \ac{ES} & \makecell{Mean Mode Swaps\\ per Episode [\si{\percent}]} & \makecell{Median Jerk\\ Norm [$\text{rad}\text{s}^{-3}$]} \\
        \hline
        \multicolumn{4}{l|}{MPC} & 0.15   & 8.00\\
        \multicolumn{4}{l|}{LSM} & 0.00   & 88.60\\
        \hline
        \textcolor{blue}{DDPM} & 5  & $\times$ & $\times$ & 0.01                  & 57.76 \\ 
        DDPM & 5  & $\times$ &  & 0.08 & 2403.78 \\ 
        DDPM & 40 &  & & 3.78    & 354.84 \\
        DDPM & 5 & & & 3.00    & 1149.10 \\
        DDIM & 5 & &  & 3.93    & 1267.84 \\
        \hline
    \end{tabularx}
    \label{tab:jerk_guidance}
\end{table}


\subsubsection{Gradient Guidance and Early Stopped Noise Injection}
\label{sec:res:guidance}
\begin{figure}[t]
\vspace{0.3cm}
  \centering
  \begin{minipage}{0.65\columnwidth}
  \centering
  \hspace{1cm}\footnotesize\textbf{Simulation}
  \end{minipage}
  \begin{minipage}{0.32\columnwidth}
  \centering
  \footnotesize
  \textbf{Hardware}
  \end{minipage}
  \includegraphics[width=\columnwidth]{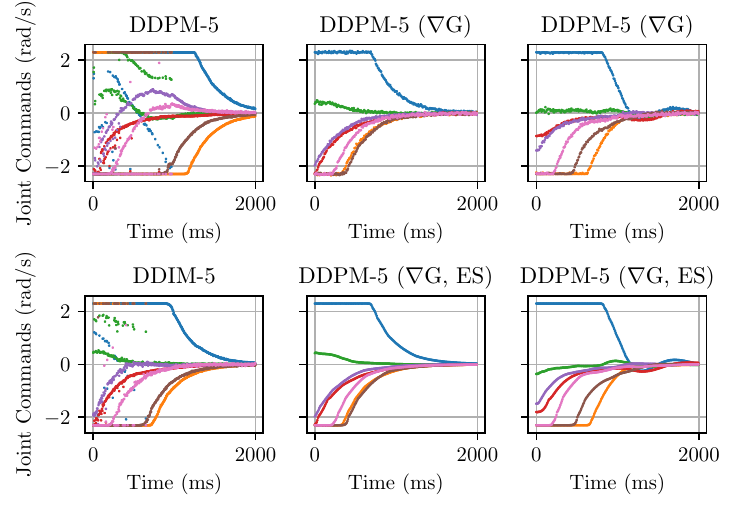}%
  \vspace{-1.2em}
  \caption{
  Exemplary closed-loop joint commands of vanilla diffusion models in simulation (left), the \ac{DDPM} with \acf{GG} in simulation (center top) and hardware (right top), the proposed \ac{DDPM} with five steps, \ac{GG}, and \acf{ES} in simulation (center bottom) and hardware (right bottom). \ac{GG} ensures consistent modes necessary for hardware deployment, while \ac{ES} smooths the commands.\looseness=-1
  }
  \label{fig:cl_denoising_comparison}
  \vspace{-3ex}
\end{figure}

We evaluate the effectiveness of \textit{i)} using \ac{GG} to remove mode swapping and \textit{ii)} reducing the noise level for later denoising steps, compared to vanilla \ac{DDPM} and \ac{DDIM}.
To this end, we compare the number of mode changes in closed loop to showcase the effectiveness of guidance and the median jerk to assess the effect of the noise schedule.
An exemplary trajectory is visualized in \Cref{fig:cl_denoising_comparison}, and quantitative metrics are reported in \Cref{tab:jerk_guidance}.
Quantitatively, guidance significantly reduces mode swaps compared to vanilla diffusion and even MPC with optimization, which underlines the effectiveness of our contribution.
While guidance shows significant jerk, early stopped noise injection reduces this by orders of magnitude and achieves smooth control with only 5 denoising steps.
In \Cref{fig:cl_denoising_comparison}, vanilla \ac{DDPM} and \ac{DDIM} with 5 denoising steps exhibit interleaved modes. 
Adding guidance provides consistency across the entire episode.
In addition, the noise schedule removes remaining jerkiness of the commands.
Notably, only policies with guidance could be deployed on hardware.\looseness=-1

\subsubsection{Feasibility Analysis}
\label{sec:res:feasibility}
\begin{figure}[t]
\vspace{0.3cm}
    \centering
      \includegraphics[width=\columnwidth, clip, trim=0cm 0.25cm 0cm 0.25cm]{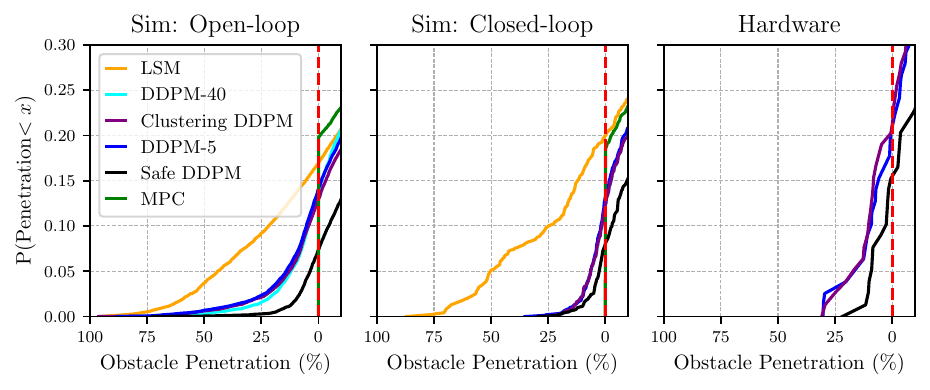}%
    \vspace{-1em}
    \caption{Constraint satisfaction of open-loop predicted action sequences (left), the resulting closed-loop trajectories in simulation (center), and closed-loop experiments on hardware (right).
    The graphs show the cumulative distribution of the penetration error of a spherical obstacle relative to its $20$ cm radius.
    All \acp{DM} are evaluated with \ac{GG} and \ac{ES}.
    }
    \label{fig:penetration_err_cdf}
    \vspace{-2ex}
\end{figure}

Thus far, tracking performance, mode consistency, and smoothness have been of significant concern.
In this section, we analyze constraint satisfaction in the presence of non-convex constraints by adding a spherical obstacle ($\diameter40$ \si{\centi\meter}) in the center of the right arm's task space.
\Cref{fig:penetration_err_cdf} shows the \textit{cumulative distribution of the obstacle penetration}.
The \ac{LSM} cannot model the non-convex constraint, thus showing poor constraint satisfaction.
\ac{DDPM} with 5 and 40 denoising steps show similar distributions in open and closed loop, with high density near the $0\%$ penetration (surface), demonstrating once more the advantage over \ac{LSM} and suggesting that five denoising steps suffice for this approximate \ac{MPC} application.\looseness=-1


\subsubsection{Computational Time}
\label{sec:res:computationaltime}
\Cref{tab:results_comp_time} shows the computational time of our approach and MPC implementation.
We achieved a $5.6\times$ speed-up on the CPU and $73\times$ on the GPU.

\begin{table}[b]
    \vspace{-3ex}
    \caption{Computational time of the proposed method compared to the \ac{MPC} and \ac{DDPM} with 40 denoising steps on the CPU or GPU.}
    \label{tab:results_comp_time}
    \begin{center}
    \begin{tabularx}{\columnwidth}{l| >{\raggedright\arraybackslash}c >{\raggedright\arraybackslash}c >{\raggedright\arraybackslash}c| >{\centering\arraybackslash}X >{\centering\arraybackslash}X >{\centering\arraybackslash}X >{\centering\arraybackslash}X}
        \hline
            \rowcolor{CaptionColor}
        Method &  $N_I$ &  \ac{GG} &  \ac{ES} & \multicolumn{2}{c}{Mean [\si{\milli\second}]} & \multicolumn{2}{c}{Perc $\mathbf{95}$th [\si{\milli\second}]} \\ 
        \rowcolor{CaptionColor} 
        & & & & \multicolumn{1}{c}{CPU} & \multicolumn{1}{c}{GPU} & \multicolumn{1}{c}{CPU} & \multicolumn{1}{c}{GPU} \\ 
        \hline
        \multicolumn{4}{l|}{MPC} & 47.54 & - & 65.63 & - \\ 
        \hline
        \textcolor{blue}{DDPM} & 5 & $\times$ & $\times$ & \textbf{9.77} & \textbf{0.85} & \textbf{11.57} & \textbf{0.90} \\ 
        DDPM & 40 & $\times$ & $\times$ & 107.18 & 3.79 & 112.76 & 3.81 \\ 
        \hline
    \end{tabularx}
    \end{center}
\end{table}


\subsection{Sampling from Diffusion Models}
So far, we have evaluated the quality of individually sampled trajectories using the \ac{DM}. 
In this section, we inspect how drawing a larger batch of samples from the \ac{DM}'s distribution can boost performance by selecting the trajectory with lowest \ac{MPC} cost, selecting only feasible trajectories, or through democratic voting via clustering.
Informative sampling reduces uncertainty and generally avoids selecting low-density modes that are either unsuccessfully denoised samples (e.g., caused by a low number of denoising steps) or low-density modes in the expert distribution that can be considered artifacts but were not removed from the dataset.

\subsubsection{Tracking Performance}
\Cref{tab:metrics_lsm_mpc_clddpm_sampling} shows how leveraging different sampling techniques improves \ac{SR} and \ac{TRT} without harming steady-state performance.
Ranking based on MPC cost, specifically $d_y(y_t^d, h(x_{N|t}))$, performs best, at a computational expense dependent on the application.
Clustering improves standard sampling without environment knowledge.\looseness=-1

\begin{table}[b]
    \vspace{-3ex}
    \caption{Closed-loop tracking performance in \textbf{non-real-time} simulation for random start and goal positions. Mean \ac{ATE}, \ac{ARE} and \ac{TRT} are computed only for successful episodes, the rate of which we report under \ac{SR}. All diffusion models are deployed using \ac{GG} and \ac{ES}.}
    \setlength\tabcolsep{4.0pt}
    \label{tab:metrics_lsm_mpc_clddpm_sampling}
    \begin{center}
    \begin{tabularx}{\columnwidth}{l| >{\raggedright\arraybackslash}c| >{\centering\arraybackslash}X >{\centering\arraybackslash}X| >{\centering\arraybackslash}X >{\centering\arraybackslash}X}
    \hline
    \rowcolor{CaptionColor} 
    Method & Sampling & \ac{ATE} [\si{\milli\meter}] & \ac{ARE} [\si{\degree}] & \ac{SR} [\si{\percent}] & \ac{TRT} [\si{\second}] \\
    \hline
    \multicolumn{2}{l|}{MPC} & \textbf{0.54}   & \textbf{0.02} & \textcolor{gray}{93.18} & 1.18 \\ 
    \multicolumn{2}{l|}{LSM} & 13.84                                        & 3.70 & 10.26 & 2.54   \\
    \hline
    \textcolor{blue}{DDPM} & - & 5.63                  & 1.29 & 95.76 & 1.14   \\
    \textcolor{blue}{DDPM} & Cluster & 5.61                   & 1.29 & 96.09 & 1.13   \\
    \textcolor{blue}{DDPM} & Cost & 5.66                  & 1.31 & \textbf{97.51}  & \textbf{1.09}   \\
    \hline
    \end{tabularx}
    \end{center}
\end{table}

\begin{table}[b]
    \vspace{-3ex}
    \centering
    \caption{Constraint satisfaction in closed-loop simulation and hardware in terms of \ac{SR} for $10\%$ and $1\%$ tolerance. Sampling multiple action sequences and choosing a safe one improves constraint satisfaction compared to naive sampling. The baseline \ac{LSM} performs poorly with non-convex constraints. The proposed approach is shown in \textcolor{blue}{blue}.}
    \vspace{0.5em}
    \label{tab:feasibility_sampling}
    \begin{tabularx}{\columnwidth}{l| >{\raggedright\arraybackslash}c| >{\centering\arraybackslash}X >{\centering\arraybackslash}X| >{\centering\arraybackslash}X >{\centering\arraybackslash}X}
        \hline
        \rowcolor{CaptionColor}
        Method & Sampling & \multicolumn{2}{c|}{Simulation} & \multicolumn{2}{c}{Hardware} \\
        \rowcolor{CaptionColor}
        & & 10\% SR & 1\%SR & 10\% SR & 1\%SR \\
        \hline
        \multicolumn{2}{l|}{MPC}                 & \textbf{100.00} & \textbf{100.00} & - & - \\
        \multicolumn{2}{l|}{LSM} & 84.20 & 80.30 & 81.25 & 75.00 \\
        \hline
        \textcolor{blue}{DDPM} &       - & 97.40 & 88.50 & 90.00 & 81.25 \\
        \textcolor{blue}{DDPM} & Cluster     & 97.70 & 88.90 & 88.75 & 80.00 \\
        \textcolor{blue}{DDPM} & Safe & 99.20  & 95.10 & \textbf{95.00} & \textbf{85.00} \\
        \hline
    \end{tabularx}
\end{table}

\subsubsection{Feasibility Analysis}
\Cref{tab:feasibility_sampling} shows feasibility metrics for sampling strategies in simulation and hardware experiments.
Clustering slightly improves performance in simulation.
However, hardware experiments are inconclusive due to the similar performance with and without clustering.
\ac{DDPM}-Safe samples $100$ trajectories with guidance and randomly picks one from the non-colliding subset, showing a clear improvement over standard \ac{DDPM}.
To avoid jerky motions, we have applied guidance to both sampling methods, thereby narrowing multi-modality for sampling strategies in closed loop and somewhat hindering performance.

\begin{table}[t]
    \caption{Computational time of the proposed sampling methods for CPU and GPU with a sample size of 100, 5 denoising steps, \ac{GG}, and \ac{ES}. The computational time increase is small compared to \Cref{tab:results_comp_time} due to parallelization (particularly on GPU). The cost sampling runs on the CPU.}
    \label{tab:results_comp_time_sampling}
    \begin{center}
    \begin{tabularx}{\columnwidth}{l| >{\raggedright\arraybackslash}c| >{\centering\arraybackslash}X >{\centering\arraybackslash}X| >{\centering\arraybackslash}X >{\centering\arraybackslash}X}
        \hline
        \rowcolor{CaptionColor}
        Method & Sampling & \multicolumn{2}{c|}{Mean [\si{\milli\second}]} & \multicolumn{2}{c}{Perc $\mathbf{95}$th [\si{\milli\second}]} \\ 
        \rowcolor{CaptionColor} 
        & & \multicolumn{1}{c}{CPU} & \multicolumn{1}{c|}{GPU} & \multicolumn{1}{c}{CPU} & \multicolumn{1}{c}{GPU} \\ 
        \hline
        \textcolor{blue}{DDPM} & Cluster & 23.602 &  \textbf{1.345} & 36.870 & \textbf{1.391} \\ 
        \textcolor{blue}{DDPM} & Safe & 28.380 & 3.305 & 45.548 & 3.525 \\ 
        \textcolor{blue}{DDPM} & Cost & 27.539 & 4.499 & 44.765 & 4.625 \\ 
        \hline
    \end{tabularx}
    \end{center}
    \vspace{-3ex}
\end{table}

\subsubsection{Computational Time}
\Cref{tab:results_comp_time_sampling} shows sampling method costs.
Modern GPUs can parallelize rollouts with negligible overhead, taking only slightly longer than naive sampling (see \Cref{tab:results_comp_time}), even with a batch size of 100.
On CPU, a batch of 100 leads to a $4\times$ increase in computational time, suitable for specific embedded applications. 

%% file: Chapters/7_Conclusions.tex
We proposed \ac{DAMPC}, a diffusion-based \ac{AMPC} formulation capable of capturing multi-modal action distributions from a nonlinear \ac{MPC} demonstrator.
By introducing specific design choices and reducing the number of denoising steps, we showed that diffusion models can be used for fast ($>$\SI{100}{\hertz}) and accurate control on real robotic systems, allowing for a speedup of more than $70 \times$ compared to the optimization-based \ac{MPC}.
By controlling a 7-\ac{DOF} manipulator that exhibits multi-modal action distributions, we showed that consistent mode selection and reducing jerk in actions are essential for successful hardware deployment, which we achieved during the denoising process using \ac{GG} and \ac{ES}.\looseness=-1

For future work, we plan to expand our results to higher-dimensional input and state spaces, e.g., vision.
Moreover, in the current setup, the required training data remains high, resulting in slow policy synthesis. A more-efficient sampling strategy during data generation could help generate new policies more efficiently.
Applying \ac{DAMPC} to embedded hardware by downsizing the \ac{MLP} or applying quantization and pruning could further open up new application domains.\looseness=-1